\documentclass[letterpaper, 10 pt, conference]{ieeeconf}
\usepackage{times}
\usepackage[pdftex]{graphicx}
\usepackage{caption}
\usepackage{subfig}

\usepackage{amsmath,amssymb,amsopn,amstext,amsfonts}
\usepackage{cancel}
\usepackage[space]{cite}
\usepackage{color}
\usepackage{mathtools}
\usepackage{algorithm}
\usepackage{algorithmicx}
\usepackage{algpseudocode}

\usepackage{bm}

\usepackage{diagbox}
\usepackage{booktabs}
\usepackage[linkcolor=black,citecolor=black,urlcolor=black,colorlinks=true]{hyperref}
\usepackage{paralist}
\makeatletter
\def\tagform@#1{\maketag@@@{\normalsize(#1)\@@italiccorr}}
\makeatother

\newcommand{\figref}[1]{Fig.~\ref{#1}}

\newcommand{\secref}[1]{Section \ref{#1}}

\graphicspath{{./Figures/}}
\DeclareGraphicsExtensions{.pdf,.png,.jpg,.eps,.svg}
\IEEEoverridecommandlockouts

\title{\LARGE \bf
Whole-Body Control With Terrain Estimation of A 6-DoF Wheeled Bipedal Robot}

\author{Cong Wen, Yunfei Li, Kexin Liu, Yixin Qiu, Xuanhong Liao, Tianyu Wang, \\
Dingchuan Liu, Tao Zhang, Ximin Lyu%
\thanks{This work was supported by the Guangdong-Hong Kong-Macao Joint Research of Science and Technology Planning Funding (Grant No. 2023A0505010019), the National Natural Science Foundation of China (Grant No. 62303495) and the Young Talent Support Project of Guangzhou Association for Science and Technology (Grant No. QT-2025-004).\textit{ (Corresponding author: Ximin Lyu) }}%
\thanks{Cong Wen, Yunfei Li, Yixin Qiu, Tianyu Wang, Dingchuan Liu, and Ximin Lyu are with the School of Intelligent Systems Engineering, Sun Yat-sen University, Guangzhou, 510275, China.}%
\thanks{Kexin liu, Xuanhong Liao is with Direct Drive Technology Limited, Dongguan, 523808, China.}%
\thanks{Tao Zhang is with the Suzhou Nuclear Power Research Institute Co., Ltd, Shenzhen, 518038, China.}
\thanks{Email:{\tt\small lvxm6@mail.sysu.edu.cn}}
}

\begin{document}
\maketitle

\begin{abstract}
Wheeled bipedal robots have garnered increasing attention in exploration and inspection. However, most research simplifies calculations by ignoring leg dynamics, thereby restricting the robot's full motion potential. Additionally, robots face challenges when traversing uneven terrain. To address the aforementioned issue, we develop a complete dynamics model and design a whole-body control framework with terrain estimation for a novel 6 degrees of freedom wheeled bipedal robot. This model incorporates the closed-loop dynamics of the robot and a ground contact model based on the estimated ground normal vector. We use a LiDAR inertial odometry framework and improved Principal Component Analysis for terrain estimation. Task controllers, including PD control law and LQR, are employed for pose control and centroidal dynamics-based balance control, respectively. Furthermore, a hierarchical optimization approach is used to solve the whole-body control problem. We validate the performance of the terrain estimation algorithm and demonstrate the algorithm's robustness and ability to traverse uneven terrain through both simulation and real-world experiments.
\end{abstract}

\section{INTRODUCTION}

\begin{figure}[t]
	\centering
        \includegraphics[width=0.65\columnwidth]{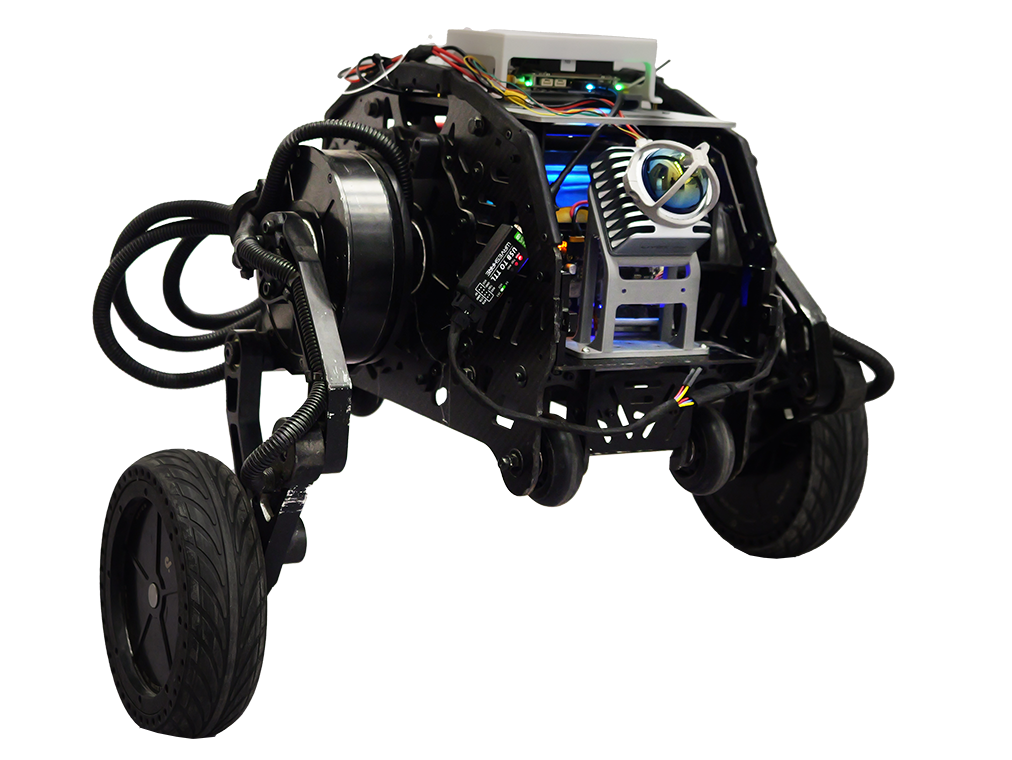}
        \caption{\textbf{DIABLO.} A 6-DoF wheeled bipedal robot composed entirely of direct drive joints. A Livox Mid-360 LiDAR installed on the robot scans the environment to acquire terrain information. For additional maneuver demos, watch the included video: https://youtu.be/E0AMqLYjEYE.}
        \label{fig:diablo}
        \vspace{-0.5cm}
\end{figure}

The wheeled bipedal robot (WBR) is a novel type of mobile robot that combines many advantages of traditional mobile robots. It retains the high mobility speeds and efficiency of wheeled robots. At the same time, its leg structure can cushion the impact when traversing uneven terrain or encountering external disturbances, and it can dynamically adjust its center of mass to adapt to varying load conditions. However, existing WBRs still have limitations when traversing uneven terrain. This study enhances WBR's adaptability on uneven terrain by integrating terrain perception data into the control system, and validates the approach on the WBR DIABLO, as shown in \figref{fig:diablo}.

In recent years, numerous WBRs with various configurations have been designed. In 2017, Boston Dynamics developed the Handle robot, showcasing remarkable mobility and capability~\cite{BD2017handle}. However, the implementation specifics of Handle are not disclosed. Ascento~\cite{klemm2019ascento}, developed by ETH Zurich, is driven by four motors, offering advantages such as a compact design and low cost. However, its mobility and exploration capabilities are also restricted by its configuration. Ollie~\cite{wang2021balance}, designed by Tencent RoboticX Lab, is a WBR with a planar parallel mechanism at two legs. The structure offers higher rigidity and stability, but it necessitates a larger volume, which limits the robot’s workspace and restricts Ollie's exploration ability in confined spaces. Taking both cost and mobility performance into comprehensive consideration, we designed a 6 degrees of freedom (DoF) WBR DIABLO with the serial mechanism at two legs in our previous work~\cite{liu2024diablo6dofwheeledbipedal}.

WBRs, a typical example of underactuated robots, are frequently modeled as a wheeled inverted pendulum (WIP) model or its variants~\cite{zhang2019system,liu2019dynamic,klemm2019ascento,chen2020underactuated,yu2023modeling,liu2024diablo6dofwheeledbipedal}. Control strategies are then developed based on these simplified models. The SR600, a 6-DoF robot designed by Zhang \textit{et al.}~\cite{zhang2019system,liu2019dynamic}, is modeled as a WIP and a PID controller is employed to achieve balance control and height control.  Likewise, Ascento~\cite{klemm2019ascento} adopts the same modeling approach and uses an LQR controller for balance control. Chen \textit{et al.}~\cite{chen2020underactuated} proposed a wheeled-spring-loaded inverted pendulum model and planned the jumping motions relying on it. Yu \textit{et al.}~\cite{yu2023modeling} designed an 8-DoF WBR and put forward a wheeled rigid body dynamics model to provide additional degrees of freedom. They utilized an MPC-based method to control the robot's pose and leg splitting. Our previous work \cite{liu2024diablo6dofwheeledbipedal} developed a second-order WIP model and performed individual dynamic analyses on the robot's rigid bodies. Additionally, a comprehensive motion controller was proposed to control the robot to complete various tasks. The above-mentioned method can achieve specific motion control for robots and has demonstrated satisfactory performance on flat ground. However, the simplified model cannot fully represent all the dynamic characteristics of WBRs. It encounters challenges in tasks that require leg cushioning, such as traversing uneven terrain. Wang~\cite{wang2024design} designed the 6-DoF WBR SKATER and proposed a hierarchical control framework.  Two control strategies were employed to adjust the robot's head roll angle, enabling the robot to achieve high-speed turning and adapt to varying terrain heights. However, this method requires a height difference between the two legs to be effective. Consequently, when traversing terrains (such as slopes) where the height difference is minimal, it may encounter challenges.

Whole-body control (WBC) is a model-based control framework designed to coordinate a robot's joint movements to accomplish multiple tasks simultaneously. By leveraging the robot's inherent redundancy, it optimally maps task-space objectives to joint-space while satisfying constraints like joint limits and contact conditions. Initially, WBC was primarily employed for controlling humanoid robots~\cite{moro2019whole}. In recent years, it has been successfully applied to WBRs. Klemm~\cite{klemm2020lqr} proposed a WBC scheme for Ascento, and the experiment demonstrated the robot's robustness in the face of external disturbances and its adaptability to varying ground heights. Ascento processes the constraints of the robot's contact with the ground by introducing contour parameters. The contour parameters depend on the ground normal vector. However, the paper does not provide a method for estimating the ground normal vector. In this paper, we propose a method for estimating the ground normal vector. When the ground information is obtained in advance, we can adopt appropriate planning and control strategies. In contrast to the passive adaptation in~\cite{wang2024design}, our method allows the robot to adapt to the terrain actively. Our experiment demonstrated that the robot exhibited a significant improvement in anti-disturbance capability and terrain adaptability. The scheme holds great potential for application in robot exploration and inspection.

We estimated ground normal vectors using point cloud data captured with LiDAR and the method of estimating point cloud surface normal vectors. Currently, point cloud surface normal vector estimation methods are categorized into traditional geometry-based methods and learning-based methods. Geometry-based methods, such as Principal Component Analysis (PCA)~\cite{hoppe1992surface} and Moving Least Squares (MLS)~\cite{levin1998approximation}, rely heavily on neighborhood size selection and depend heavily on the user's experience. Learning-based methods, such as PCPNet~\cite{guerrero2018pcpnet}, DeepFit\cite{ben2020deepfit}, AdaFit\cite{zhu2021adafit}, and TRFit~\cite{liu2023trfit}, significantly improve surface normal vector prediction, especially in challenging areas like edges and corners. However, they require substantial computational resources and exhibit slower processing speeds~\cite{zhang2020mixed}. To ensure real-time and stable applicability, we adopted an improved adaptive optimal neighborhoods PCA method~\cite{zou2019Anewmethod} for terrain estimation.

The main contributions of the paper are as follows:
\begin{itemize}
    \item Derivation of the complete 3D dynamics model of a closed-loop WBR(\secref{sec:modeling}).
    \item  Proposal of an online terrain estimation algorithm based on LiDAR(\secref{sec:estimation}).
    \item Development of a WBC framework integrated with terrain estimation based on our robot(\secref{sec:control}).
\end{itemize}

\section{MODELING}\label{sec:modeling}
As illustrated in \figref{fig:generalized_coordinates}(a), we define the inertia frame \textbf{I}, the floating base frame \textbf{B}, and the contact frame $\textbf{C}$ ($\textbf{C}_l$ is the contact frame of the left wheel and $\textbf{C}_r$ is the contact frame of the right wheel). In this paper, the red, green, and blue arrows represent the $x$, $y$, and $z$ axes of a coordinate system, respectively. Additionally, the bold lowercase letters are used to denote vectors and the bold uppercase letters to denote matrices.

\subsection{Whole-body Dynamics}

\begin{figure}[t]
    \centering
     \includegraphics[width=0.82\columnwidth]{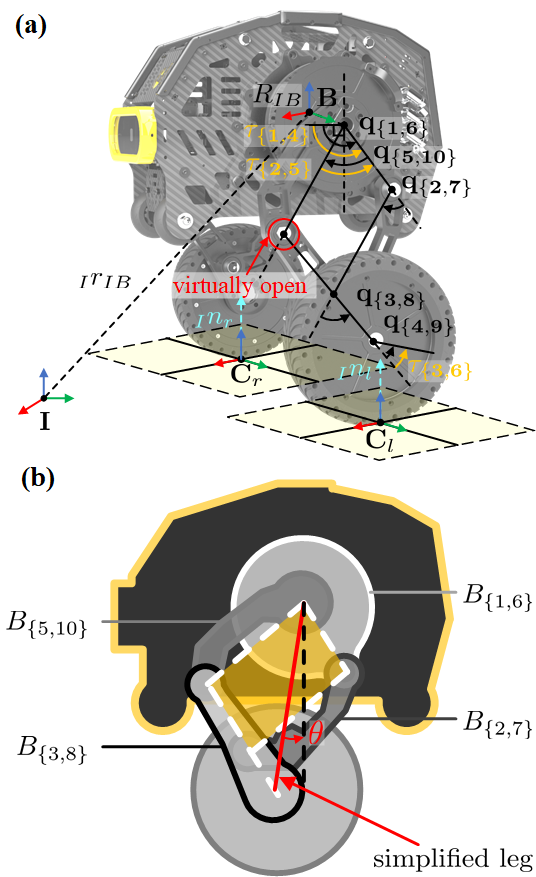}
    \caption{\textbf{Robot coordinate systems, generalized coordinates, actuated torques, and leg structure.} (a) $q_{\{left,right\}}$ and $\tau_{\{left,right\}}$ respectively represent the generalized coordinates and the actuated torques on both sides of the robot. $_I\bm{n}_l$ and $_I\bm{n}_r$ are ground normal vectors at the contact points between the wheel and the ground. (b) $B_{\{left,right\}}$ represent the rigid bodies on both sides of the robot.  The red line represents a simplified version of the leg linkage. The pendulum angle $\theta$ is the angle between the simplified leg and the vertical direction.} 
    \label{fig:generalized_coordinates}
    \vspace{-0.5cm}
\end{figure}

A commonly used strategy for formulating the equation of motion (EoM) of a closed-loop system is to first virtually cut a passive joint in the kinematic chain~\cite{khalil2002modeling,featherstone2014rigid},  thereby generating a spanning tree of the system. The loop-closure constraints are then applied to the EoM of the spanning tree.

As shown in \figref{fig:generalized_coordinates}(a), we define generalized coordinates $\bm{q}$, and actuated torques $\bm{\tau}_a$ of the spanning tree as follows:
\begin{equation}
    \bm{q} = \begin{bmatrix}
    \bm{q}_b\\\bm{q}_j
    \end{bmatrix},
    \bm{q}_b = \begin{bmatrix}
    _I\bm{r}_{IB}\\\mathbf{R}_{IB}
    \end{bmatrix},
    \bm{q}_j = \begin{bmatrix}
    q_1\\ \vdots \\q_{10}
    \end{bmatrix},
    \bm{\tau}_a = \begin{bmatrix}
    \tau_1\\ \vdots \\\tau_{6}
    \end{bmatrix},
\end{equation}
where  $\bm{q}_b\in \mathbb{R}^{3}\times SO(3)$ represents the un-actuated base coordinates and $\bm{q}_j\in \mathbb{R}^{10}$ denotes the joint coordinates. $_I\bm{r}_{IB}$  and $\mathbf{R}_{IB}$ are the translation and the rotation of the base.

The EoM of the spanning tree can be formulated as follows
\begin{equation}\label{eq:open loop dynamics}
    \mathbf{H}(\bm{q})\bm{\dot u} + \mathbf{C}(\bm{q},\bm{u})= \mathbf{S}^T\bm{\tau}_a + \bm{\tau}_{gc}~,
\end{equation}
where $\mathbf{H}\in\mathbb{R}^{16\times16}$ denotes the generalized inertia matrix, $\mathbf{C}\in\mathbb{R}^{16}$ represents the generalized bias force, which accounts for Coriolis forces, centripetal forces, and gravity terms. $\bm{u}\in  \mathbb{R}^{16} $ and $\bm{\dot u}\in  \mathbb{R}^{16} $ denote the sets of generalized velocities and accelerations, respectively.
\begin{equation}
     u=\begin{bmatrix}
        _I\bm{v}_{IB}&_I\bm{\omega}_{IB}&\dot q_1&\cdots&\dot q_{10}
    \end{bmatrix}^T.
\end{equation}
$_I\bm{v}_{IB}$ and $_I\bm{\omega}_{IB}$ are the linear and angular velocities, $\mathbf{S} \in \mathbb{R}^{16\times 6}$ is the selection matrix and $\bm{\tau}_a \in  \mathbb{R}^{6}$ represents the torque of actuated joints. $\bm{\tau}_{gc}\in \mathbb{R}^{16}$ represents the ground contact force applied in joint space.

As shown in \figref{fig:generalized_coordinates}(b), each leg of the robot features a parallelogram mechanism composed of four rigid bodies, which imposes explicit motion constraints on the robot system
\begin{equation}\label{eq:parallel}
\begin{aligned}
    q_5 &= q_2 = -q_3,\\
    q_{10} &= q_7 = -q_8.
\end{aligned}
\end{equation}
Let $\bm{y}\in \mathbb{R}^{3} \times SO(3)\times \mathbb{R}^{6}$ represent a vector of independent position variables for the closed loop system, which defines $\bm{q}$ uniquely. We define $\bm{y}$ as follows:
\begin{equation}
    \bm{y}=\begin{bmatrix}
        _I\bm{r}_{IB}&\mathbf{R}_{IB}&q_1&q_5&q_4&q_6&q_{10}&q_9
    \end{bmatrix}^T,
\end{equation}
and provide a loop closure function that satisfies the following
\begin{equation}\label{eq: parallelogram motion constraint}
    \bm{q} = \gamma(\bm{y}).
\end{equation}
Differentiating this equation gives
\begin{equation}
    \mathbf{G}=\frac{\partial{\gamma}}{\partial{\bm{y}}},
\end{equation}
We define $\bm{\tau}_c$ as the constraint force that limits leg movement. By introducing this force into \eqref{eq:open loop dynamics}, the EoM for the closed-loop system is
\begin{equation}\label{eq: closed dynamics}
      \mathbf{H}(\bm{q})\bm{\dot u} + \mathbf{C}(\bm{q},\bm{u})= \mathbf{S}^T\bm{\tau}_a + \bm{\tau}_{gc}+\bm{\tau}_c~. 
\end{equation}
According to Jourdain's principle of virtual power\cite{featherstone2014rigid}. $\bm{\tau}_c$ has the following property
\begin{equation}
      \mathbf{G}^T\bm{\tau}_c = 0.
\end{equation}
Therefore, by premultiplying $\mathbf{G}^T$ on \eqref{eq: closed dynamics} to eliminate $\bm{\tau}_c$, we obtain an alternative form of the EoM for the closed system
\begin{equation}\label{eq: closed-loop EoM}
    \mathbf{H}_y\bm{\dot u}_y + \mathbf{C}_y = \mathbf{G}^T\mathbf{S}^T\bm{\tau}_a + \mathbf{G}^T\bm{\tau}_{gc},
\end{equation}
where $\mathbf{H}_y=\mathbf{G}^T\mathbf{H}\mathbf{G}\in\mathbb{R}^{12\times12}$, $\mathbf{C}_y = \mathbf{G}^T\mathbf{C}\in\mathbb{R}^{12}$, and we obtain a new set of generalized coordinates $\bm{y}$, velocities $\bm{u}_y\in\mathbb{R}^{12}$ and accelerations $\bm{\dot u}_y\in\mathbb{R}^{12}$
\begin{equation}
    u_y=\begin{bmatrix}
        _I\bm{v}_{IB}&_I\bm{\omega}_{IB}&\dot q_1&\dot q_5&\dot q_4&\dot q_6&\dot q_{10}&\dot q_9
    \end{bmatrix}^T.
\end{equation}

\begin{figure*}[th]
	\centering
        \includegraphics[width=\textwidth]{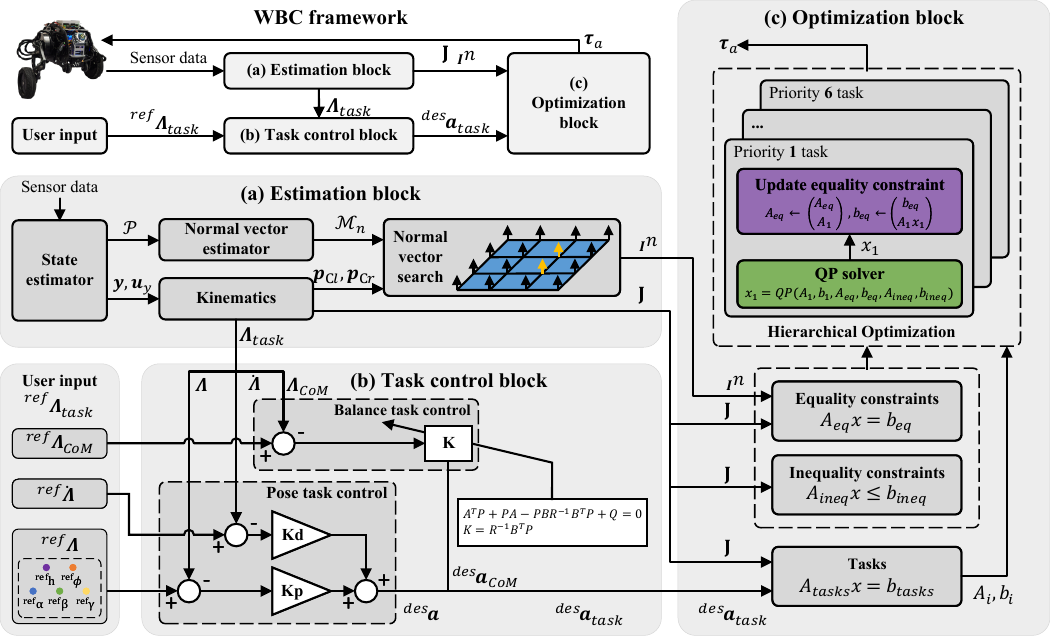}
        \caption{\textbf{WBC framework}  }
        \label{fig:WBC_framework}
        \vspace{-0.5cm}
\end{figure*}
\subsection{Ground Contact}\label{sec:ground contact}
We use the method described in~\cite{klemm2020lqr} to calculate $\bm{\tau}_{gc}$. To prevent relative motion between the contact points and the ground, the acceleration in the x and z directions of the contact frame, $_C\bm{a}^{x,z}_{IC}\in\mathbb{R}^2$ (the superscript denotes the components of a vector in specific directions) is constrained to zero, and $\bm{\tau}_{gc}$ can be formulated as follow
\begin{equation}\label{eq:ground contact}
    \bm{\tau}_{gc}=(\underbrace{({}_{C}\mathbf{J}^{x,z}_{IC})^T+  ({}_{C}\mathbf{J}^{y}_{IC})^T\mathbf{C}_F}_{\mathbf{J}_{gc}})\bm{F}_C,
\end{equation}
where ${}_C\mathbf{J}^{x}_{IC}$, ${}_C\mathbf{J}^{y}_{IC}$, ${}_C\mathbf{J}^{z}_{IC}\in \mathbb{R}^{2\times12}$ are the contact Jacobian matrices of two sides. $\bm{F}_C\in \mathbb{R}^4$ is the rolling constraint forces, $\mathbf{C}_F\in \mathbb{R}^{2\times 4}$ represents a velocity-dependent friction coefficient matrix. The calculations of ${}_C\mathbf{J}_{IC}$ and $\mathbf{C}_F$ are related to the ground normal vector $_I\bm{n}$.
By combining \eqref{eq: closed-loop EoM} and \eqref{eq:ground contact}, we get the final EoM
\begin{equation}\label{eq:final EoM}
\begin{aligned}
       & \mathbf{H}_y\bm{\dot u}_y + \mathbf{C}_y = \mathbf{G}^T\mathbf{S}^T\bm{\tau}_a + \mathbf{G}^T\mathbf{J}_{gc}\bm{F}_C~,\\
       & _C\bm{a}^{x,z}_{IC}={}_C\mathbf{\dot J}^{x,z}_{IC}\bm{u}_y+{}_C\mathbf{J}^{x,z}_{IC}\bm{\dot u}_y=0.
\end{aligned}
\end{equation}

\section{CONTROL WITH TERRAIN ESTIMATION}\label{sec:control}

As illustrated in \figref{fig:WBC_framework}, our WBC framework is divided into three blocks: (a) Estimation Block: This block processes sensor data and calculates the robot's task state $\bm{\Lambda}_{task}$, Jacobian matrix $\mathbf{J}$, and the ground normal vector $_I\bm{n}$ on the contact point of the two legs. (b) Task Control Block: The task controller takes the user-defined reference task state $^{ref}\bm{\Lambda}_{task}$ and the estimated task state $\bm{\Lambda}_{task}$ to compute the desired task acceleration $^{des}\bm{a}_{task}$, which is then passed to the optimization block. (c) Optimization block: This block receives $\mathbf{J}$, $_I\bm{n}$ and $^{des}\bm{a}_{task}$ from the previous two blocks. They are used to formulate the optimization problem and solve for the actuated torques $\bm{\tau}_a$, which are then used as inputs to the robot's actuators.

\subsection{Estimation Block}\label{sec:estimation}

As shown in \figref{fig:WBC_framework}(a), the sensor data was fed to the state estimator module to get generalized coordinates $\bm{y}$, velocities $\bm{u}_y$ and point cloud set $\mathcal{P}$. In this paper, we use the EKF to estimate $\bm{y}$ and $\bm{u}_y$, and a computationally efficient LiDAR-inertial odometry framework FAST-LIO~\cite{xu2021fast}~\cite{xu2022fast} to get $\mathcal{P}$. Then, $\bm{y}$ and $\bm{u}_y$ were fed to the kinematics module to calculate $\mathbf{J}$, $\bm{\Lambda}_{task}$ and contact position $\bm{p}_{Cl}$, $\bm{p}_{Cr}$ of two wheels. By using the improved PCA, the normal vector estimator calculates the normal vectors and stores them in a global map $\mathcal{M}_n$. After that, the normal vector search module utilized $\bm{p}_{Cl}$, $\bm{p}_{Cr}$ to search the ground normal vector $_I\bm{n}$ in $\mathcal{M}_n$.

FAST-LIO processes raw point cloud data and IMU data to compute a global point cloud frame, which is then merged into the $\mathcal{P}$. PCA is commonly employed to estimate the normal vectors on $\mathcal{P}$. This method delivers precise estimations on smooth surfaces with notable efficiency. However, the outcomes are remarkably sensitive to the selected neighborhood size. Inadequate choices in this aspect can introduce instability during the estimation process.

To address this issue, we adopted an improved PCA method~\cite{zou2019Anewmethod} that minimizes an entropy function to determine the optimal neighborhood size k for each vertex
\begin{equation}
    k = \mathop{\arg\min}\limits_{k}(-\eta_{k1}\ln(\eta_{k1})-\eta_{k2}\ln(\eta_{k2})-\eta_{k3}\ln(\eta_{k3})),
\end{equation}
\begin{equation}
    \eta_{ki} = \lambda_{ki}/\sum^3_{i=1}\lambda_{ki},
\end{equation}
where $\lambda_{k1}$, $\lambda_{k2}$, and $\lambda_{k3}$ are three eigenvalues of the covariance matrix for neighborhood size k. This method balances accuracy, speed, and robustness without too much effort in manually selecting neighborhood sizes.

After determining the neighborhood size, we compute the covariance matrix of all vertices within the neighborhood of the current point cloud vertex. The eigenvector corresponding to the smallest eigenvalue of the matrix is taken as the normal vector of the current vertex.

\subsection{Task Control Block}\label{sec: task control block}
As shown in \figref{fig:WBC_framework}(b), the user provides the reference state $^{ref}\bm{\Lambda}_{task}$, and the estimation block provides the estimated states $\bm{\Lambda}_{task}$ to the task controllers. The robot's motion tasks are divided into a balance task and pose tasks.  The balance task, responsible for CoM stabilization, employs an LQR controller to compute the desired CoM acceleration $^{des}\bm{a}_{CoM}$. In parallel, each pose task is controlled by an individual PD controller, which computes the corresponding desired acceleration $^{des}\bm{a}$. We reorder $^{des}\bm{a}_{CoM}$ and $^{des}\bm{a}$ based on task priority to obtain the $^{des}\bm{a}_{task}$.

\subsubsection{Pose task control}
Pose tasks include the split angle task, height task, and head orientation task. 

The robot is equipped with 6 identical actuators at the hips, knees, and wheels, enabling agile motions such as crouching, jumping, and leg splitting, similar to a roller skater~\cite{liu2024diablo6dofwheeledbipedal}. The split angle $\phi$ is defined as the difference between the pendulum angles of the left and right legs. Notably, to ensure stable turning and prevent slippage during locomotion, it is necessary to maintain a zero wheel separation $\bm{d}_w = 0$, which indirectly limits the split angle. 

To describe this constraint and other pose tasks,  we define the local control frame \textbf{N}, similar to the approach in~\cite{klemm2020lqr}, as shown in \figref{fig:local_control_frame}. The origin of frame \textbf{N} is at the midpoint of contact points $C_l$ and $C_r$. In frame \textbf{N}, the x-axis aligns with the direction of the robot's head, and the z-axis aligns with the normal vector of the horizontal plane.

In the horizontal plane, $\bm{d}_w$ can be formulated as
\begin{equation}
    \bm{d}_w = {}_N\bm{r}_{NCl}^x-{}_N\bm{r}_{NCr}^x~.
\end{equation}
The robot's height also defined $h$ in the frame \textbf{N} ,
\begin{equation}
    h = {}_N\bm{r}_{NB}^z,
\end{equation}
and we use Euler angles, calculated in the ZYX order, to determine the orientation of the robot's head. $\gamma,\beta,\alpha$ in the ZYX order for the base. $\gamma,\beta,\alpha$ represent the yaw, pitch, and roll angles, respectively. Then, we define the pose task state as 
\begin{equation}\label{eq: task state}
    \bm{\Lambda} = \begin{bmatrix}
        \phi&h&\alpha&\beta&\gamma
    \end{bmatrix}^T\in \mathbb{R}^5.
\end{equation}
The control objectives for pose tasks are decoupled, and we use the PD control law to control them. n is the index in $\bm{\Lambda}$, A pose task with index n can be expressed as
\begin{equation}
    {}^{des} a_n = K_{pn}({}^{ref}\bm{\Lambda}_n-\bm{\Lambda}_n)+K_{dn}({}^{ref}\bm{\dot \Lambda}_n-\bm{\dot \Lambda}_n),
\end{equation}
where $K_{pn}$ and $K_{dn}$ are proportional gain and derivative gain, respectively. The parameter tuning procedure is as follows: first, adjust $K_{pn}$ to ensure the system can track the given input. Then, set $K_{dn}$ to $\sqrt{K_{pn}}$, if oscillations occur, reduce $K_{pn}$. Finally, define the desired accelerations of pose tasks
\begin{equation}
    {}^{des} \bm{a} = \begin{bmatrix}
        {}^{des} a_1&\cdots&{}^{des} a_5
    \end{bmatrix}^T\in \mathbb{R}^5.
\end{equation}

\subsubsection{Balance task control}
\begin{figure}[t]
	\centering
        \includegraphics[width=\columnwidth]{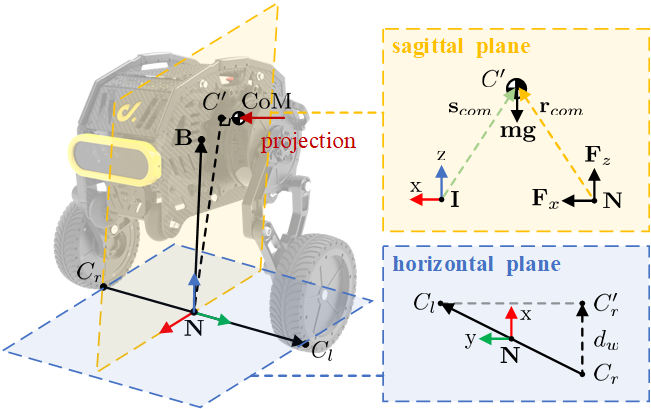}
        \caption{\textbf{Local control frame.} The sagittal plane and the horizontal plane of the robot. $C'$ is the orthographic projection of CoM onto the sagittal plane. $C'_r$ is the contact point of the right wheel when the split angle equals zero.}
        \label{fig:local_control_frame}
        \vspace{-0.5cm}
\end{figure}

We adopt LQR as the balance control strategy. By simplifying the robot as a single rigid body and orthogonally projecting its CoM onto the point $C'$ on the sagittal plane, we develop a centroidal dynamics model, as illustrated in \figref{fig:local_control_frame}.

We define $\boldsymbol{k}_{CoM} \in \mathbb{R}^{3}$ as the centroidal momentum of the robot, which includes the centroidal linear momentum $\boldsymbol{p}_{CoM}\in \mathbb{R}^{2}$, the centroidal angular momentum $\boldsymbol{N}_{CoM}\in \mathbb{R}$ and the contact force $\bm{F}_{NC} \in \mathbb{R}^{2}$, all in the sagittal plane
\begin{equation}
    \boldsymbol{k}_{CoM}=
\begin{bmatrix}
     {\boldsymbol{p}}_{CoM} \\
     {\boldsymbol{N}}_{CoM}
\end{bmatrix},
\bm{F}_{NC} =  
\begin{bmatrix}
     F_x \\
     F_z
\end{bmatrix}.
\end{equation}

According to the Newton–Euler method, 
\begin{equation}\label{eq:centroidal_momentum}
\begin{aligned}
\dot{\boldsymbol{k}}_{CoM}=
\begin{bmatrix}
    m\bm{\ddot s}_{CoM}\\
    I\bm{\dot \omega}_{CoM}
\end{bmatrix}=
\bm{W}_g+\bm{W}_{gc},\\
    \bm{W}_g = 
    \begin{bmatrix}
    -m\mathbf{g}\\
    0\end{bmatrix},
    \bm{W}_{gc}= \begin{bmatrix}\bm{F}_{NC} \\
    -\bm{r}_{CoM}\times \bm{F}_{NC}\end{bmatrix},
    \end{aligned}
\end{equation}
where $\bm{W}_g$ denotes the gravity wrench, $\bm{W}_{gc}$ denotes the ground contact wrench, and $\bm{s}_{CoM}$, $\bm{\omega}_{CoM}$ and $\bm{r}_{CoM}$ denote the absolute position, angular velocity, and relative position of the CoM for the origin of frame \textbf{N}, respectively. In the task space, we have:
\begin{equation}
\begin{aligned}
    \bm{r}_{CoM} &= {}_I\bm{r}_{NC'},\\
    \bm{s}_{CoM} &= {}_I\bm{r}_{IC'}.
\end{aligned}
\end{equation}

To maintain balance, the robot must satisfy the following constraints: zero acceleration along the Z-axis and zero angular acceleration.

\begin{equation}\label{eq:balance_constraint}
\begin{aligned}
    m\bm{\ddot s}_{CoM}^z&=mg+F_z=0,\\
    I\bm{\dot \omega}_{CoM}&=-r_{CoM}^zF_x+r_{CoM}^xF_z=0.
\end{aligned}    
\end{equation}

Considering constraints in \eqref{eq:balance_constraint}, we derive the following state-space representation
\begin{equation}\label{eq:com controller}
    \begin{bmatrix}
        \bm{\dot r}_{CoM}^x\\ \bm{\ddot r}_{CoM}^x \\ \bm{\dot s}_{CoM}^x \\ \bm{\ddot s}_{CoM}^x
    \end{bmatrix}=
    \begin{bmatrix}
        0&1&0&0\\
        0&0&0&0\\
        0&0&0&1\\
        \bm{g}/\bm{r}_{CoM}^z&0&0&0
    \end{bmatrix}
    \begin{bmatrix}
        \bm{r}_{CoM}^x\\ \bm{\dot r}_{CoM}^x \\ \bm{s}_{CoM}^x \\ \bm{\dot s}_{CoM}^x
    \end{bmatrix}+
    \begin{bmatrix}
        0\\1\\0\\0
    \end{bmatrix}
    \bm{\ddot r}_{CoM}^x.
\end{equation}

By solving the Ricatti equation, we obtain the gain matrix $\mathbf{K}$, and then we obtain the desired relative acceleration of the CoM
\begin{equation}
\begin{aligned}
         ^{des}\bm{\ddot r}_{CoM}^x &= -K({}^{ref}\bm{\Lambda}_{CoM}-\bm{\Lambda}_{CoM}),\\
    \bm{\Lambda}_{CoM} &= \begin{bmatrix}
        \bm{r}_{CoM}^x& \bm{\dot r}_{CoM}^x & \bm{s}_{CoM}^x &\bm{\dot s}_{CoM}^x
    \end{bmatrix}.
\end{aligned}
\end{equation}
Additionally, we define the desired acceleration of the balance task
\begin{equation}
    {}^{des} \bm{a}_{CoM} = ^{des}\bm{\ddot r}_{CoM}^x.
\end{equation}

\subsection{Optimization Block}

The WBC executes tasks by coordinating all joint motions, and we formulate it as a constrained optimization problem. The optimization variable is defined as
\begin{equation}
    \bm{x} = \begin{bmatrix}
        \bm{\dot u}_y^T & \bm{F}_C^T & \bm{\tau}_a^T
    \end{bmatrix}^T\in \mathbb{R}^{22}.
\end{equation}

To ensure dynamic feasibility, the optimization must satisfy the EoM \eqref{eq:final EoM}, which are reformulated as the following linear equality constraints
\begin{equation}\label{eq:dynamics constriants}
\begin{aligned}
&\qquad \mathbf{A}_{eq}\bm{x}=\bm{b}_{eq},\\
        \mathbf{A}_{eq}&=
    \begin{bmatrix}
    \mathbf{H}_y&  -\mathbf{G}^T\mathbf{J}_{gc} &-\mathbf{G}^T\mathbf{S}^T\\
    {}_C\mathbf{J}^{x,z}_{IC}&0&0
    \end{bmatrix},\\
     \bm{b}_{eq}&=
    \begin{bmatrix}
    -\mathbf{C}_y\\
    -{}_C\mathbf{\dot J}^{x,z}_{IC}\bm{u}_y
    \end{bmatrix}.
\end{aligned}
\end{equation}

Since generalized accelerations $\bm{\dot u}_y$ are part of the optimization variables, we use task controllers to describe all tasks at the acceleration level.  The desired task acceleration $^{des}\bm{a}_{task}\in\mathbb{R}^6$  is constructed by reordering $^{des}\bm{a}$ and $^{des}\bm{a}_{CoM}$ according to the task priority:  height, pitch angle, balance, roll angle, split angle and yaw angle. For each task with priority i, the corresponding Jacobian matrix $\mathbf{J}_i$ maps the generalized accelerations to the task space 
\begin{equation}\label{eq:jacobian map}
    \mathbf{J}_i\bm{\dot u}_y + \mathbf{\dot J}_i \bm{u}_y = {}^{des}a_{task,i},
\end{equation}
and \eqref{eq:jacobian map} can be formulated as least square problem with 
\begin{equation}\label{eq:LS}
    \mathbf{A}_i = \begin{bmatrix}
        \mathbf{J}_i & 0 & 0
    \end{bmatrix},\bm{b}_{i}={}^{des}a_{task,i}-\mathbf{\dot J}_i \bm{u}_y.
\end{equation}
We then stack all 6 tasks according to their priority order into the matrix $\mathbf{A}_{task}$ and the vector $\bm{b}_{task}$,
\begin{equation}\label{eq:task stack}
\begin{aligned}
    \mathbf{A}_{task}&=\begin{bmatrix}
        \mathbf{A}_1&\dots&\mathbf{A}_6
    \end{bmatrix}\\
    \bm{b}_{task}&=\begin{bmatrix}
        \bm{b}_1&\dots&\bm{b}_6
    \end{bmatrix}.
\end{aligned}
\end{equation}

We solve the problem using a sequence of constrained quadratic programs (QP). As shown in \figref{fig:WBC_framework}, to ensure tasks are executed according to their priority, we adopt a hierarchical optimization approach. The QP problem with priority i can be written as
\begin{equation}
\begin{aligned}
\min_{x} \frac{1}{2} \| \mathbf{A}_i\bm{x}-\bm{b}_{i} \|^2,\\
\text{s.t. } 
     \begin{cases}
         \mathbf{A}_{eq,i}\bm{x} = \bm{b}_{eq,i}\\
         \mathbf{A}_{ineq}\bm{x} \leq \bm{b}_{ineq}
     \end{cases},
\end{aligned}
\end{equation}
where $\mathbf{A}_{eq,i}$, $\bm{b}_{eq,i}$ denotes the equality constraints. $\mathbf{A}_{ineq}$, $\bm{b}_{ineq}$ denote the inequality constraints, which enforce the limits on the actuation torques. In each optimization iteration, the result from the previous higher-priority level is used as an equality constraint for the subsequent lower-priority level. After the final iteration, the actuated torques $\bm{\tau}_a$, which are part of the optimization variables, will be input to the robot's actuators.

\begin{figure}[t]
    \centering

    \subfloat[]{
        \includegraphics[width=\columnwidth]{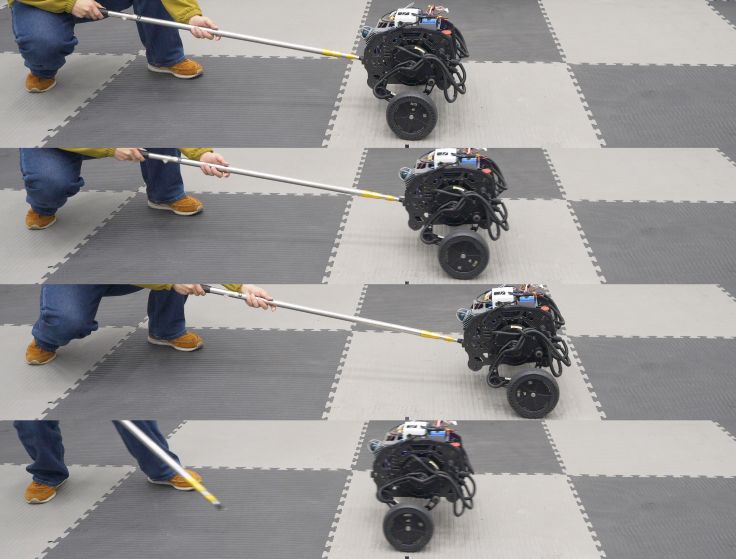}
    }\\
    \vspace{-0.4cm} 
    \subfloat[]{
        \includegraphics[width=\columnwidth]{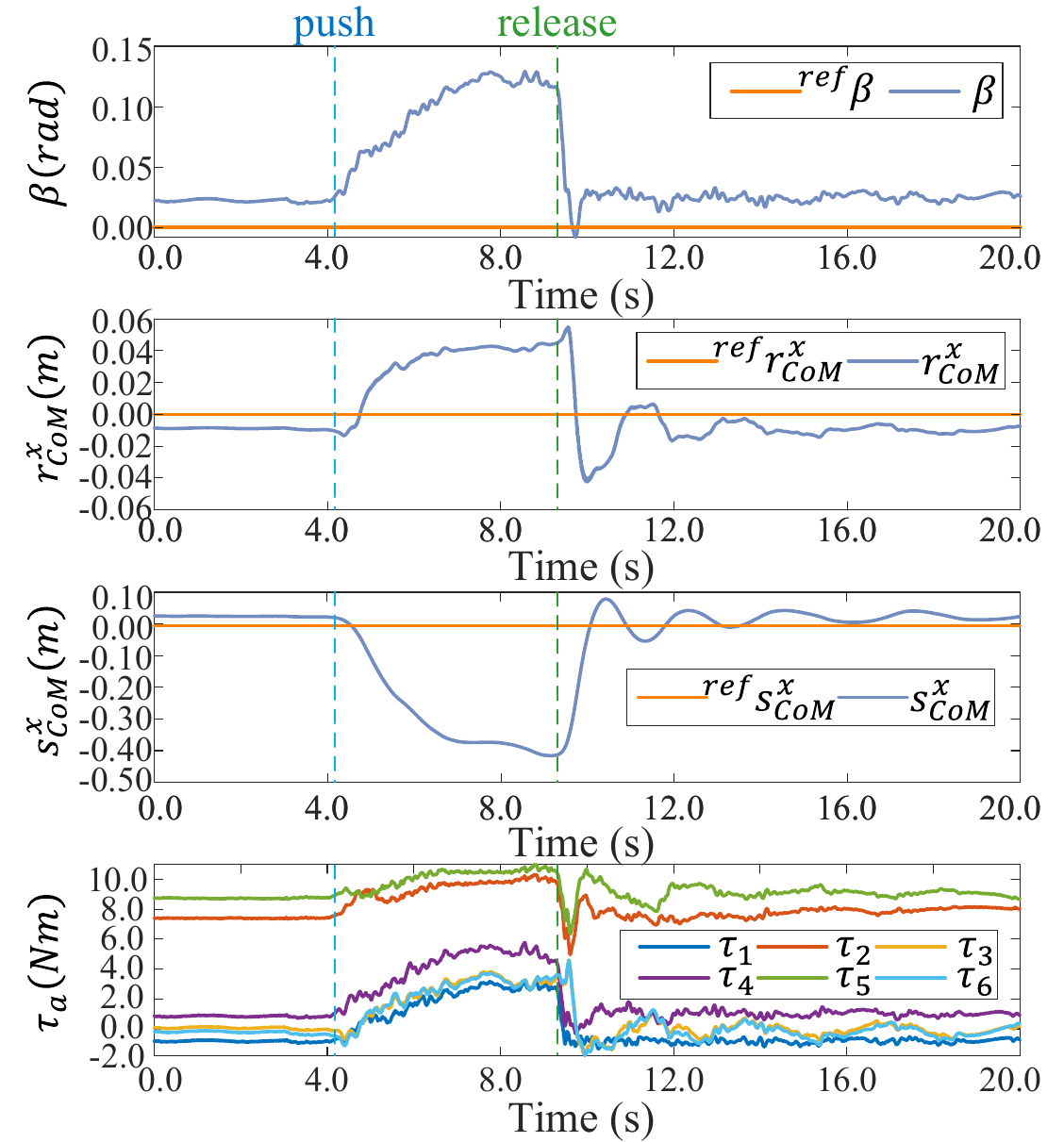}
        }
    \caption{\textbf{Disturbance recovery experiment.} (a) The robot is disturbed and recovers. (b) The evolution of the robot's pitch angle $\beta$,  CoM deviation distance $\bm{r}_{CoM}^x$, absolute position $\bm{s}_{CoM}^x$, and actuated torques $\bm{\tau}_a$.}
    \label{fig:exp1}
    \vspace{-0.5cm}
\end{figure}

\begin{figure}[t]
    \centering
    \subfloat[]{
        \includegraphics[width=\columnwidth]{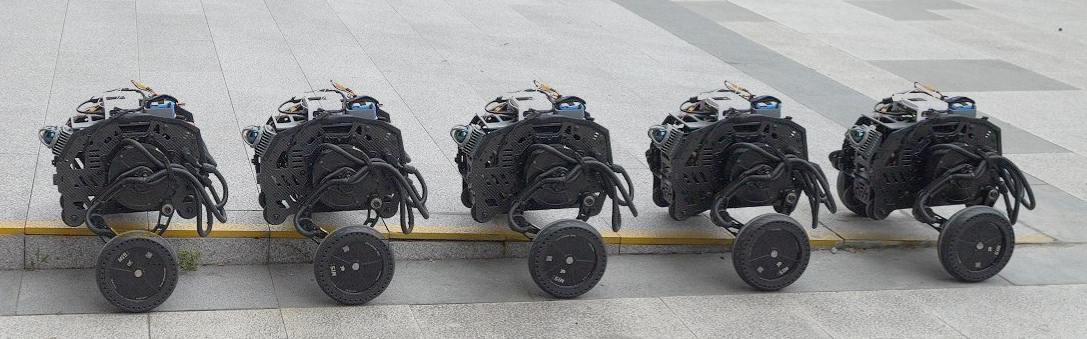}
    }\\
    \vspace{-0.5cm} 
    \subfloat[]{
        \includegraphics[width=\columnwidth]{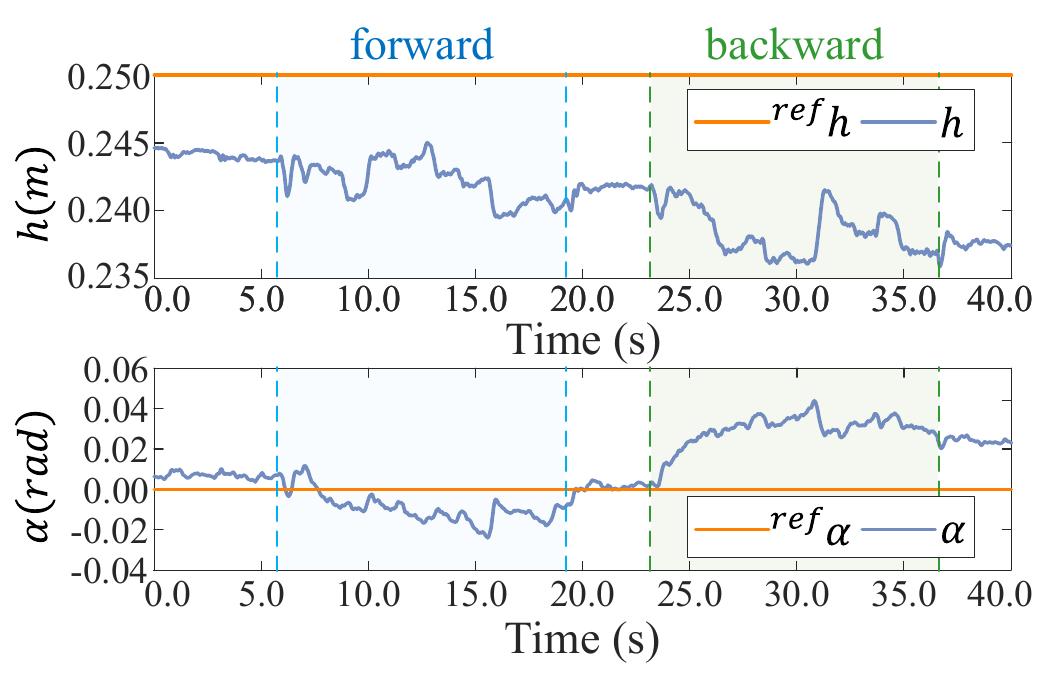}
        }
    \caption{\textbf{Adaptation to varying ground heights experiment.} (a) Snapshots of the experiment. (b) The evolution of the robot's height $h$ and roll angle $\alpha$.}
    \vspace{-0.3cm}
    \label{fig:exp2}
\end{figure}

\begin{figure}
    \centering
        \includegraphics[width=\columnwidth]{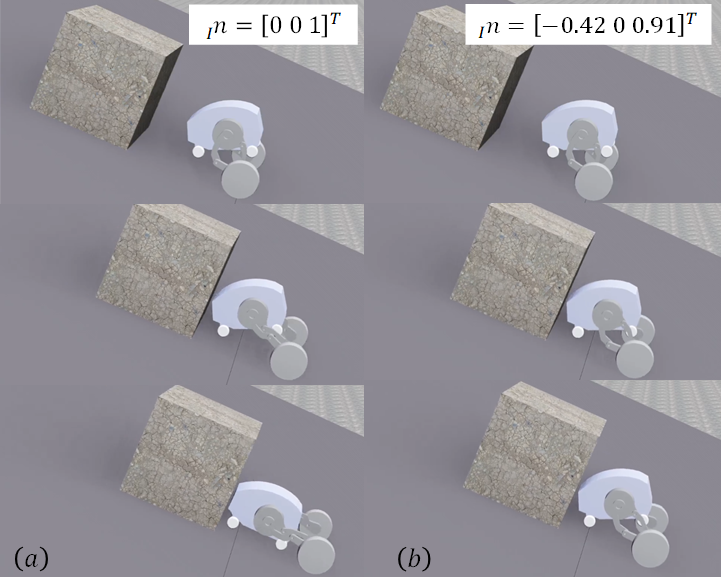}
        \caption{\textbf{Slope impact comparative experiment.} (a) Without the true normal vector (knocked over) (b) With true normal vector (maintains balance).}
        \label{fig:exp3_1}
        \vspace{-0.5cm}
\end{figure}

\begin{figure}[t]
    \centering

    \subfloat[]{
        \includegraphics[width=\columnwidth]{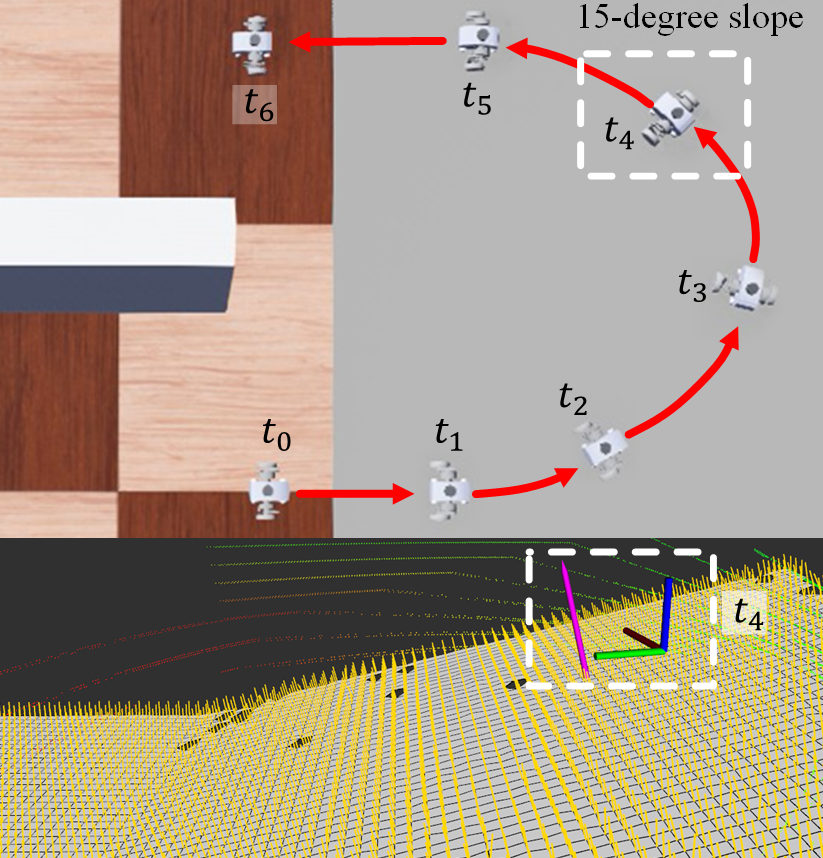}
    }\\
    \vspace{-0.4cm} 
    \subfloat[]{
        \includegraphics[width=\columnwidth]{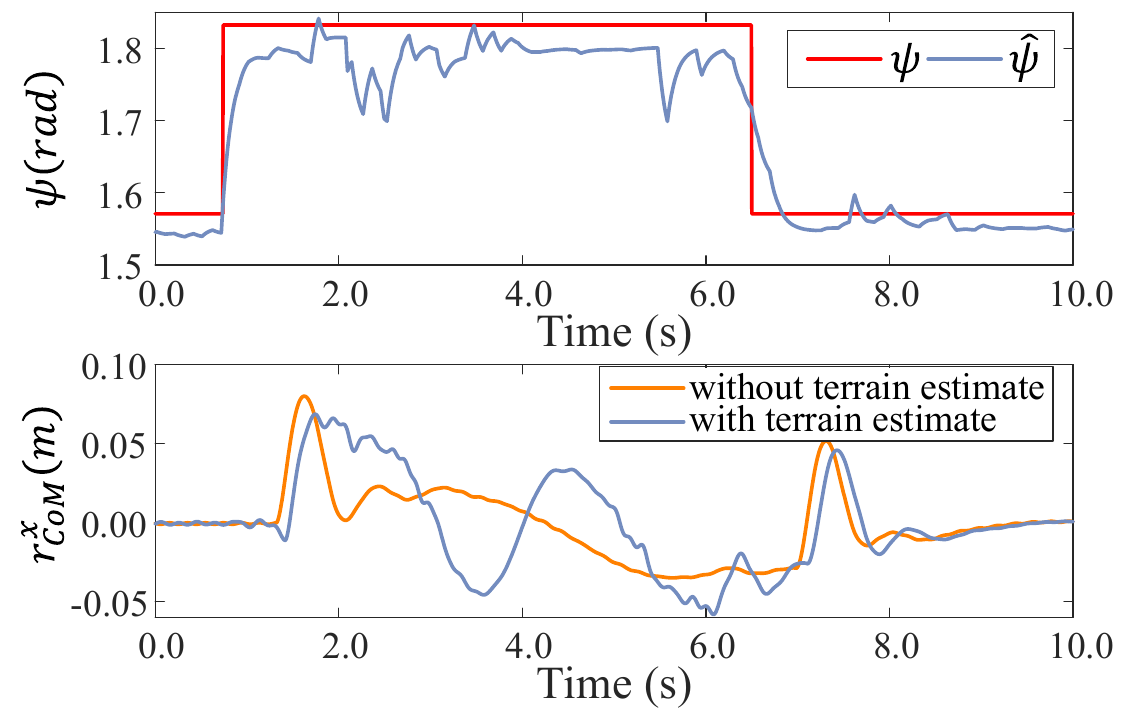}
        }
    
    \caption{\textbf{Slope U-turn maneuvering experiment} (a) Snapshots at different time points and terrain estimation visualization at time $t_4$. (b) The evolution of the estimated inclination angle $\hat\psi$ and robot's CoM deviation distance $\bm{r}_{CoM}^x$. $\psi$ is the real inclination angle.}
    \vspace{-0.5cm}
    \label{fig:exp3_2}
\end{figure}

\section{EXPERIMENTS}\label{sec:experiments}

We experimentally validated the WBC algorithm both in the real world and in simulation. In the real-world implementation, we developed the control system on the ROS2 control framework, employed the Pinocchio dynamics library, and used qpOASES for optimization. For simulations, we used Webots, MATLAB, and the Spatial v2 dynamics library.

\subsection{Disturbance recovery experiment}

We designed the disturbance recovery experiment in the real world to verify our WBC algorithm's ability to maintain balance, as shown in \figref{fig:exp1}(a), we use a rod to push the robot and gradually increase the pushing force before releasing it. 

As shown in \figref{fig:exp1}(b), during the process, the pitch $\beta$ variation was 0.128 rad (7.34$^{\circ}$), and hip torques $\tau_{\{1,4\}}$ increased in response to disturbances, preventing excessive pitch angle change. The robot's absolute position $\bm{s}_{CoM}^x$ was pushed back 0.41 m, and the CoM deviation $\bm{r}_{CoM}^x$ reached 0.06 m. As the applied force increased, the knee torques $\tau_{\{2,5\}}$ and wheel torques $\tau_{\{3,6\}}$ also increased, making it progressively more difficult to push the robot. After release, the robot returned to its original position within 1 second. The experiment demonstrates the algorithm's robustness and the ability to maintain the robot's posture under external disturbances, which is beneficial for perception stability.

\subsection{Adaptation to varying ground heights experiment}

We designed the experiment in the real world to validate the adaptability of our WBC algorithm to varying ground heights. As shown in \figref{fig:exp2}(a). In this experiment, the robot moved forward and backward on the asymmetric ground support. At the highest point, the right contact point is 0.1 m higher than the left.

As shown in \figref{fig:exp2}(b), we input the reference height state $^{ref}h$ to 0.25 m, as indicated by the orange line. Throughout the process, the robot adjusted the contraction and extension of its left and right legs to maintain posture and prevent tipping. The robot's height $h$ variation remained within 1 cm and its roll angle $\alpha$ deviation to both the left and right remained within 0.04 rad (2.3°). During forward movement, the robot's average height was 0.2426 m with a standard deviation (SD) of 0.0016 m, with an average roll angle of -0.4882° (SD: 0.3965°). In backward movement, the average height was 0.2406 m (SD: 0.0017 m), with an average roll angle of 1.5507° (SD: 0.4708°). Compared to the reference, the height errors were only 3.0\% and 3.8\% during forward and backward movement, respectively. This experiment demonstrated the WBC algorithm's traversal and adaptability capabilities on uneven terrain.

\subsection{Terrain Estimation simulation experiments}

\subsubsection{Slope impact comparative experiment}
We designed the experiment in the simulation to demonstrate the improvement in terrain adaptability. We place the robot on a 25$^{\circ}$ slope and release a 9 kg block (the block has zero friction) from 0.55 m above its inclined position. As shown in \figref{fig:exp3_1}, we compare the robot’s response to the impact when given the horizontal normal vector, as shown in \figref{fig:exp3_1}(a), and the true normal vector, as shown in \figref{fig:exp3_1}(b). The results show that the robot using the true normal vector remains balanced after the impact, while the one using the horizontal normal vector is knocked over.

\subsubsection{Slope U-turn maneuvering experiment}
We designed the experiment in the simulation, as shown in \figref{fig:exp3_2}(a), the robot moves up the 15$^{\circ}$ slope at a speed of 1.5m/s and makes a U-turn, then returns to flat ground. At a specific moment $t_4$ during the robot's descent, we visualized the terrain estimation algorithm. The point cloud data detected by the LiDAR is displayed using a rainbow color. We used a grid map to present the global normal vector map $\mathcal{M}_n$, where small yellow lines represent the estimated normal vectors for each grid. The tricolored coordinate axis in the image indicates the robot's current position and orientation, while the purple arrow indicates the position and direction of the currently estimated normal vector.

To enable the robot to detect changes in the ground normal vector earlier, we estimate the normal vector at a point 0.9 m ahead of the current position. Additionally, to mitigate the impact of sudden terrain changes on the robot, a low-pass filter is applied to smooth the perceived normal vector. 

As shown in \figref{fig:exp3_2}(b), we calculate the inclination angle $\hat\psi$ using the estimated ground normal vector to make the normal vector estimation results more intuitive. The estimated $\hat\psi$ aligns well with terrain contours, and the average estimation of the slope is 12.3$^{\circ}$ with an estimated error of 2.7$^{\circ}$. Our terrain estimation demonstrates accuracy. 

We also compared the WBC controller with and without terrain estimation. While entering the slope, the robot without terrain normal vector estimation experienced a CoM deviation of 0.08 m. Similarly, the CoM deviation without estimation reached 0.051 m upon exiting the slope. In contrast, with terrain normal vector estimation, the robot's CoM deviation was 0.068 m when entering the slope and 0.045 m when exiting. Compared to the robot without estimation, the CoM deviation was reduced by 15\% when entering the slope and by 12\% when exiting the slope. These results indicate that the robot experiences less CoM deviation when the robot exits the current terrain, demonstrating improved terrain adaptability.

\section{CONCLUSION AND FUTURE WORK}\label{sec:conclusion}

In this paper, we designed a WBC framework for our closed-loop WBR, derived the complete dynamics model, and proposed an online terrain estimation to estimate the ground normal vector. Furthermore, we utilized task controllers to control tasks and performed hierarchical optimization to solve the WBC problem. In the real-world experiments, the robot demonstrated disturbance rejection, terrain adaptability, and head stabilization. In the slope impact comparative simulation experiment, we showcased the improvement in terrain adaptability achieved by incorporating the ground normal vector. In the slope U-turn maneuvering simulation experiments, our terrain estimation system exhibited high accuracy.

In the future, we plan to expand on our current research to explore terrain-based planning and control strategies.

\bibliography{refs} 
\end{document}